\documentclass[conference]{IEEEtran}
\IEEEoverridecommandlockouts
\usepackage{cite}
\usepackage{amsmath,amssymb,amsfonts}
\usepackage{algorithmic}
\usepackage{graphicx}
\usepackage{textcomp}
\usepackage{xcolor}
\usepackage{paralist}
\usepackage{hyperref}
\usepackage{multirow}

\def\BibTeX{{\rm B\kern-.05em{\sc i\kern-.025em b}\kern-.08em
    T\kern-.1667em\lower.7ex\hbox{E}\kern-.125emX}}
\begin{document}

\title{Are Concept Drift Detectors Reliable Alarming Systems? - A Comparative Study}

\author{\IEEEauthorblockN{1\textsuperscript{st} Lorena Poenaru-Olaru}
\IEEEauthorblockA{\textit{Software Engineering} \\
\textit{TU Delft}\\
Delft, Netherlands \\
L.Poenaru-Olaru@tudelft.nl}
\and
\IEEEauthorblockN{2\textsuperscript{nd} Luis Cruz}
\IEEEauthorblockA{\textit{Software Engineering} \\
\textit{TU Delft}\\
Delft, Netherlands \\
L.Cruz@tudelft.nl}
\and
\IEEEauthorblockN{3\textsuperscript{rd} Arie van Deursen}
\IEEEauthorblockA{\textit{Software Engineering} \\
\textit{TU Delft}\\
Delft, Netherlands \\
arie.vandeursen@tudelft.nl}
\and
\IEEEauthorblockN{4\textsuperscript{th} Jan S. Rellermeyer}
\IEEEauthorblockA{\textit{Dependable and Scalable Software Systems} \\
\textit{Leibniz University Hannover}\\
Hanover, Germany \\
rellermeyer@vss.uni-hannover.de}
}

\maketitle

\begin{abstract}

As machine learning models increasingly replace traditional business logic in the production system, their lifecycle management is becoming a significant concern. Once deployed into production, the machine learning models are constantly evaluated on new streaming data. Given the continuous data flow, shifting data, also known as concept drift, is ubiquitous in such settings. Concept drift usually impacts the performance of machine learning models, thus, identifying the moment when concept drift occurs is required. Concept drift is identified through concept drift detectors. In this work, we assess the reliability of concept drift detectors to identify drift in time by exploring how late are they reporting drifts and how many false alarms are they signaling. We compare the performance of the most popular drift detectors belonging to two different concept drift detector groups, error rate-based detectors and data distribution-based detectors. We assess their performance on both synthetic and real-world data. In the case of synthetic data, we investigate the performance of detectors to identify two types of concept drift, abrupt and gradual. Our findings aim to help practitioners understand which drift detector should be employed in different situations and, to achieve this, we share a list of the most important observations made throughout this study, which can serve as guidelines for practical usage. Furthermore, based on our empirical results, we analyze the suitability of each concept drift detection group to be used as alarming system.

\end{abstract}


\section{Introduction}

Predictive algorithms, such as classification algorithms using Machine Learning (ML) on Big Data have seen a significant growth in interest and plenty of real-world applications have been proposed. Examples of those applications are fault detection~\cite{faultdetection}, anomaly detection~\cite{anomalydetection}, weather prediction~\cite{weatherprediction}, or credit risk prediction \cite{creditriskds}, where different ML models are constantly evaluated on streaming data. Generally, due to the continuous data flow, data streams are more prone to changes in data distributions over time and, thereby, to concept drift.

Concept drift is a significant threat to the performance of ML models over time. ML models are created by training an ML algorithm on a certain amount of available data, which we are referring to as reference data. The ML algorithms work under the assumption that the data distribution used to evaluate the model is similar to the data distribution of the reference data. However, this assumption does not hold when considering data streams since the evaluation (testing) data is constantly evolving over time due to uncontrollable factors~\cite{conceptdriftapplications}. Therefore, this raises a substantial issue with regards to preserving the performance of ML models over time.

Knowing beforehand when concept drift occurs could help data scientists to take appropriate measures in advance to prevent its effects on the ML model's performance~\cite{learningUnderConceptDrift}. Thus, special drift algorithms called \emph{concept drift detectors} were proposed to identify the moment when concept drift occurs. They can be used as an alarming system that notifies users about expected model performance degradation. Consequentially, it is important for these drift detectors to be precise when reporting the moment of data shift.

Several studies have identified two major concept drift detectors categories, the \textit{error rate-based} drift detectors and the \textit{data distribution-based} drift detectors~\cite{BAYRAM2022108632},~\cite{learningUnderConceptDrift}. The error rate-based drift detectors identify drift by monitoring the error rate of a trained model on new evaluation data batches. They are always paired with the classification algorithm used to train the model. Since they continuously compute the error rate, these detectors assume that labels are available immediately, which makes them \textit{label-dependent drift detectors}. The data distribution-based drift detectors identify drifts by assessing the similarity between the distribution of the reference data and the evaluation data. There is currently no general similarity metric used uniformly among all studies. Since their drift detection mechanism solely relies on density functions of training and testing data, they are \textit{label-independent drift detectors}. In real-world settings, the data distribution-based detectors are favored over the error rate-based detectors since immediately obtaining labels can be expensive or even impossible~\cite{Gama2014ASO}. However, recently some techniques were developed to adapt error rate-based detectors for unsupervised and semi-supervised settings~\cite{BAYRAM2022108632}. Previous comparative studies \cite{largescalecomparison}, \cite{novelhybridpaircomparative}, \cite{acomparativestudyonconceptdriftdetection} focused on analyzing only the error rate-based detectors. Thus, our study is the first to compare the aforementioned two categories of drift detectors. Furthermore, we are the first to assess the precision of detectors in terms of latency and false-positive rate from the perspective of monitoring Big Data ML applications in production and to provide guidelines for practitioners. Thereby, we contribute in the following directions:

\begin{compactenum}[1)]
    \item We assess both the data distribution-based detectors and the error-rate based detectors in terms false alarms, miss-detection rate and drift detection latency on both synthetic and real-world data.
    \item We explore different similarity metrics of data distribution-based detectors and find that, in some cases, other similarity distances are more suitable than the widely used KL  Divergence~\cite{kdqtrees},~\cite{pcacd}.
    \item We share the open source implementation of the data distribution-based detectors employed in this study, which was not previously available. Furthermore, our work is reproducible and available on GitHub.
    \item We evaluate the error rate-based detectors paired with three recent and popular classifiers, such as AdaBoost~\cite{adaboost}, XGBoost~\cite{xgboost} and LightGBM~\cite{lightgbm}, as well as commonly used classifiers~\cite{largescalecomparison,novelhybridpaircomparative,acomparativestudyonconceptdriftdetection}, i.e., Naive Bayes and Hoeffding Trees.
    \item We provide some major observations of detector-dataset compatibility, which aim to serve as guidelines for practitioners who want to include drift detectors in their data stream monitoring process.
    
\end{compactenum}

\section{Background and Related Work}
\subsection{Concept Drift General Knowledge}
\begin{figure}
\centering
\includegraphics[width=0.43\textwidth]{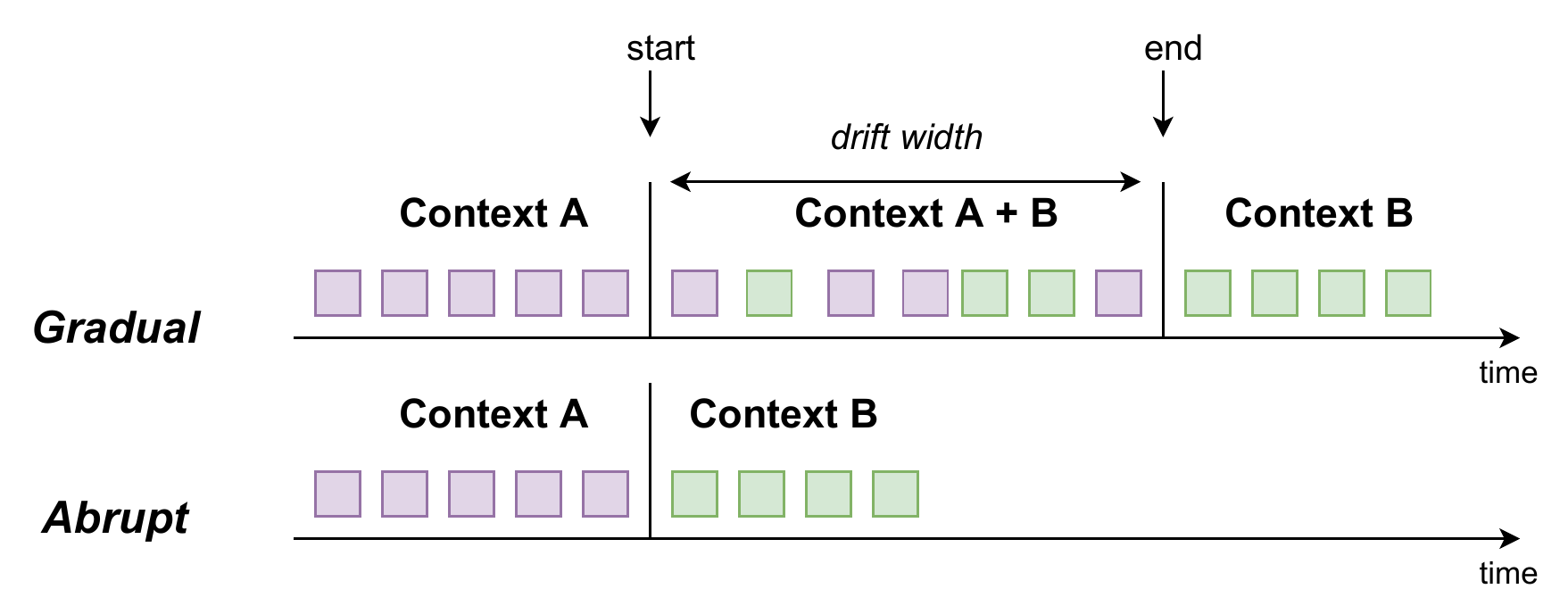}
\caption{Gradual vs abrupt drift duration.}
\label{figure:gradual_abrupt_drift}
\end{figure}

The term of concept drift, also known as data shift or data drift, was originally used in data streams to describe changes in data distributions over time~\cite{Gama2014ASO}. The most common types of concept drift are \textit{abrupt drift} and \textit{gradual drift}~\cite{acomparativestudyonconceptdriftdetection,largescalecomparison,novelhybridpaircomparative}. The key difference between the two types of drift is the duration. In case of abrupt drift, there is a sudden change in the feature behavior, while in case of gradual change, the features are changing completely after a transition period, as it can also be observed in Fig.~\ref{figure:gradual_abrupt_drift}. The transition period between the moment when gradual drift starts and the moment it ends is referred to as \textit{drift width}.

\subsection{Concept Drift Detectors}
Plenty of attention has been paid to developing techniques that are able to detect concept drift as part of data stream monitoring~\cite{learningUnderConceptDrift}. This section presents drift detectors belonging to both \textit{error rate-based (ERB)} drift detectors and \textit{data distribution-based (DDB)} drift detectors.

Out of the ERB drift detectors, the most popular drift detector is \textit{Drift Detection Method \textbf{(DDM)}}~\cite{ddm}, which uses statistical tests to identify significant changes in error rate. An improved version of DDM is \textit{Early Drift Detection Method \textbf{(EDDM)}}~\cite{eddm}, which, additionally, verifies the distance between error rates when identifying drifts. Another popular ERB drift detector is \textit{Adaptive Windowing \textbf{(ADWIN)}}~\cite{adwin}, a window based technique to store recent samples. The decrease in mean of the stored samples is monitored to detect drift. The \textit{Drift Detection Method based on the Hoeffding’s inequality carried with A-test \textbf{(HDDM\_A)} and W-test \textbf{(HDDM\_W)}}~\cite{hddm} methods rely on tracking the moving average and Hoeffding's inequality to determine the significance of the change.  Other examples of ERB drift detectors are \textit{FW-DDM}~\cite{fwddm}, \textit{EWMA chart drift detector}~\cite{ewma} and \textit{RDDM}~\cite{rddm}.

Within the (DDB) detectors we distinguish between detectors employing \textit{statistical tests} and detectors using \textit{similarity metrics}. The most popular example of the former category is the \textit{Equal Density Estimation \textbf{(EDE)}} detector~\cite{ede}, which identifies drift based on a non-parametric statistical tests. The null hypothesis of the tests assumes the similarity of two data distributions and its rejection signals a drift. The most commonly employed DDB detector relying on similarity metrics is \textit{quad-trees which scale with the size (k) and dimensionality (d) of the data \textbf{(kdqTrees)}}~\cite{kdqtrees}. This technique uses bootstrapping to determine the highest discrepancy between the reference (training) data and subsamples of the reference data in order to compute a critical region. Thereafter, the similarity between the distribution of the new data and the reference data, assessed by the critical region, is used to detect drift. For this technique the similarity metric used is KL Divergence. However, other studies consider different similarity metrics to measure similarities between distributions~\cite{rd}.  Therefore, there is no general similarity metric used in DDB drift detectors and no available study about different metrics suitability in concept drift detection. Furthermore, recent studies suggest that extracting the distributions of the projected features obtained through Principal Component Analysis (PCA) instead of raw features is more suitable for high dimensional datasets~\cite{pcacd} and could significantly improve drift detection. Other DDB drift detectors are SyncStream~\cite{synctream} or RD~\cite{rd}.

\subsection{Datasets for Concept Drift Detectors Evaluation}

When comparing concept drift detectors, most studies \cite{acomparativestudyonconceptdriftdetection, largescalecomparison, novelhybridpaircomparative} are relying on synthetic datasets, usually generated through the MOA Framework~\cite{moa}. The reason for this is that the moment when the concept drift occurs could be fixed through data generation.

Evaluating concept drift detectors on real-world data is most of the times impractical given that the exact moment when a drift occurs is unknown. However, the study of Webb et al. \cite{elect2airlinesdrift} identifies the moment of drift occurrence for two \textit{real-world datasets}, Electricity (ELECT2)~\cite{elect2} and Airlines~\cite{moa}. 

The \textit{ELECT2} datasets, contains samples from Australian New South Wales Electricity Market collected every five minutes over a period of approximately two years. The main prediction problem of ELECT2 is determining whether prices are going up or down based on demand and supply features. In this dataset there is a sudden drift on the 2\textsuperscript{nd} of May when wholesale electricity sales between the Australian Capital Territory, New South Wales, South Australia and Victoria was allowed~\cite{elect2airlinesdrift}. The effect of this concept drift could be observed on three attributes of the dataset, which were constant until that date, but started fluctuating afterwards.

The \textit{Airlines dataset}, contains samples corresponding to details of multiple flights collected over a period of four weeks. The main prediction problem is determining whether flights are going to be delayed or on-time. Within this dataset, there is a significant concept drift occurring during the weekend flights (starting from Friday until Sunday) compared to the week days. This drift can be observed especially on the first two weeks of collected data~\cite{elect2airlinesdrift}.


\section{Evaluation Methodology}
The main goal of this paper is to evaluate the ability to detect drift in time of both error-rate based (ERB) drift detectors and the data distribution-based (DDB) drift detectors under different conditions. This can be summarized in the following research questions:
\begin{compactenum}[I]
    \item[\textbf{RQ1:}] How do state-of-the-art drift detectors compare in their ability to detect abrupt and gradual drift under ideal circumstances?
    \item[\textbf{RQ2:}] How do state-of-the-art drift detectors perform detecting abrupt and gradual drift in the presence of noise and imbalanced data?
    \item[\textbf{RQ3:}] To what extent does the performance of drift detectors on controlled drift position data generalize to real-world data?
\end{compactenum}

\subsection{Data}
\subsubsection{Employed Datasets}

In order achieve our goal, we need to know precisely when the drift occurs. Thereby, in our evaluation we include both synthetic data, where the concept drift can be fixed through the data generation process, and real-world data for which we know the moment when the concept drift occurs~\cite{elect2airlinesdrift}. Our study exploits three \textit{synthetic datasets}, namely \textit{SEA}, \textit{AGRAW1} and \textit{AGRAW2}, and two real-world datasets for which the moment of concept drift occurrence is known and marked through the findings of Webb et al.~\cite{elect2airlinesdrift}, namely \textit{Electricity (ELECT2)} and \textit{Airlines}.

We generated synthetic data through two data generators, namely SEA~\cite{sea} and Agrawal~\cite{agrawal} available in the MOA framework. It needs to be mentioned that MOA was solely employed to generate the data, not to perform the evaluation. The former generates three attributes containing numerical features ranging from 0 to 1 and is frequently used in the concept drift detection literature~\cite{acomparativestudyonconceptdriftdetection,largescalecomparison},~\cite{novelhybridpaircomparative}. The latter creates three categorical attributes and six numerical attributes, which correspond to loan-related data. Agrawal generator was created through the process of database mining, in which significant patterns were extracted from large scale industrial data sets and used to generate synthetic data samples. We generated two datasets with the Agrawal generator, \textit{AGRAW1} and \textit{AGRAW2}. Although both AGRAW1 and AGRAW2 were generated using the same generator, they are two different datasets, which consider different forms of evaluation when classifying the samples into the two classes. For all the synthetic datasets, we generated data under ideal conditions, in which no noise was added and the two classes are balanced and also non-ideal conditions, with 10\% and 20\% noise or imbalanced classes, where the imbalance ratio is 1:2. This is the highest imbalance ratio for which the detectors were able to identify any drift. The scope of the non-ideal conditions is to assess the robustness of the detectors against events that could occurs in real-world scenarios. Furthermore, we generate data for both abrupt and gradual drift. We consider different drift widths, namely [500, 1000, 5000, 10000, 20000] samples. For instance, from the moment the gradual concept drift starts, there are 500 samples until it ends and the features are changing their behavior completely. Each dataset is generated using 10 random seeds to avoid bias in our experiments. We further assess the ability of drift detectors to identify drift on two real-world datasets.

We purposely include datasets containing solely numerical features, SEA and ELECT2, as well as datasets containing both numerical and categorical features, AGRAW1, AGRAW2 and Airlines. In general, categorical features pose problems for ML classifiers, since the ML algorithms usually require numerical values. The most commonly used technique to overcome this issue is one-hot encoding, which converts categorical data into binary vectors. Therefore, we employ this technique in our experiments for the datasets containing a mixture of numerical and categorical features.

After ensuring that all datasets contain solely numerical values, the next step is the data scaling. Two of the datasets considered, SEA and ELECT2, include data scaled between 0 and 1. In order to ensure experimental consistency, we also scale the values for remaining datasets, AGRAW1, AGRAW2 and Airlines. Data scaling was performed using the Min-Max scaler implementation provided in the Python-scikit learn package \footnote{scikit-learn version 1.0.2 \url{https://scikit-learn.org/stable/modules/generated/sklearn.preprocessing.MinMaxScaler.html}}. The reason for choosing this scaler is that it does not make any assumption regarding the distribution of the data following a particular pattern.
\subsubsection{Data Setup}
\begin{figure}
    \centering
    \includegraphics[width=0.43\textwidth]{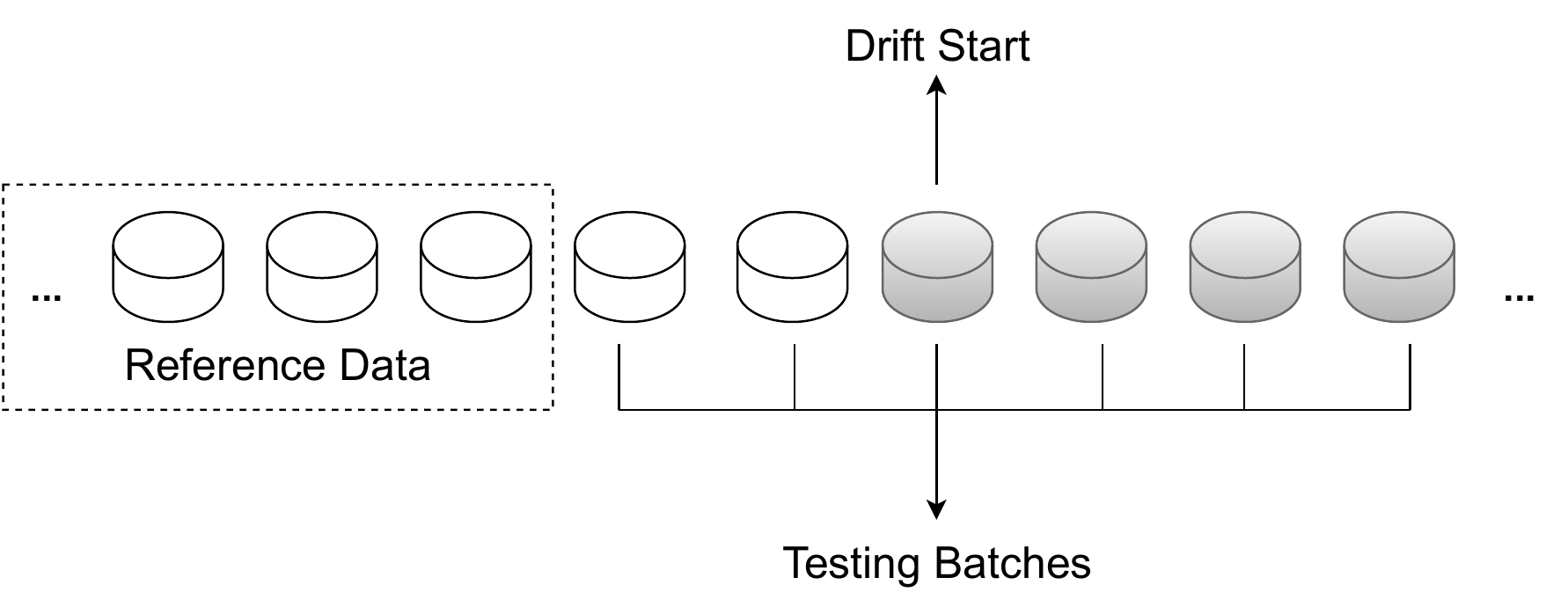}
    \caption{Data stream setup.}
    \label{figure:setup}
\end{figure}

In our experiments we process each dataset as a data stream in which the first part is the reference data and the second part is the testing data. The testing data is divided into equal testing batches. The reason behind this is that the reference data is used to train the ML model which is going to be periodically tested on the new upcoming data. In all cases we ensure that the drift occurs during the testing phase, such that we simulate a deployed ML model which needs to be tested on shifted data. For each new testing batch, a drift detector is employed to determine whether the data has shifted. A detailed representation of our setup is shown in Fig. \ref{figure:setup}. In all our experiments, detectors that signal the drift before the testing batch containing the actual concept drift is a false alarm. In the same manner, signaling the drift after the testing batch containing the drift increases the latency. In case of ERB detectors, the reference data is used in order to train the ML classifiers, which are paired with the concept drift detectors. In case of some DDB detectors, the reference data is used to compute a threshold, which is employed to assess the similarity between the new data and the old data. Furthermore, we solely use the reference data to fit the scaler and then we apply it on each testing batch.


In the SEA, AGRAW1 and AGRAW2 datasets, the drift start is fixed during the data synthesizing process such that the first two testing batches do not contain drift, while the others include drift. 

In case of the real-world data, we initially defined the prediction problem. In case of ELECT2, the prediction problem is weekly predicting whether prices are going up or down. The reference data is composed of the initial part of the data stream, namely data collected between the 07\textsuperscript{th} of May 1996 and the 15\textsuperscript{th} of April 1997. Each testing batch is composed of one week of data. The drift starts in the testing batch containing the 2\textsuperscript{nd} of May 1997. In terms of the Airline dataset, the prediction problem is daily predicting delayed flights. Since the first week of data has missing records corresponding to Monday and Tuesday, we solely consider the second week, for which we have complete data from Monday until Sunday. The reference data is represented by samples from Monday and Tuesday, while the other week days are testing batches.  The drift starts on the testing batch corresponding to Friday and lasts until the end of the week.

\subsection{Implementation Decisions in Drift Detectors}
When selecting the detectors used for evaluation, our major selection criteria is the publicly available implementations. However, one barrier we encountered was the implementation unavailability of the DDB detectors, for which only mathematical proofs were provided. Thus, we implemented three popular such detectors.

In terms of \textbf{ERB} drift detectors we employ DDM, EDDM, ADWIN, HDDM\_A and HDDM\_W, using the implementations provided in the Python package scikit-multiflow\footnote{scikit-multiflow version  0.5.3. available here: \url{https://scikit-multiflow.github.io} - this package was chosen based on its popularity}. These detectors rely on the error rate and, thus, they need to be paired with classifiers. For this study, we use the following classifiers \textit{Naive Bayes}, \textit{Hoeffding Trees}, \textit{AdaBoost}, \textit{XGBoost} and \textit{LightGBM}, which were used either in previous drift detection comparative studies,~\cite{acomparativestudyonconceptdriftdetection}, ~\cite{largescalecomparison}, \cite{novelhybridpaircomparative} or in data stream classification \cite{xgboostdatastream}, \cite{adaboostdatastream}. The classifiers are not retrained after a drift is signaled since the purpose of the experiment is solely to identify how fast the first drift can be captured, not to evaluate the situations of multiple drifts. Therefore, the reference data is also not changed after a drift is signaled. For each detector we employed the best hyperparameters configuration.

When it comes to \textbf{DDB} drift detectors, we employ the statistical test-based detector EDE and two similarity metric-based detectors, namely kdqTrees and PCA-kdq. We implemented both EDE and kdqTrees according to the original papers \cite{ede}, \cite{kdqtrees}. In case of EDE, we employed two non-parametric statistical tests, namely Kolmogorov-Smirnov and Mann Whitney. When it comes to kdqTrees, the original implementation included KL Divergence as similarity metric. However, in our work, we experimented with seven similarity metrics corresponding to seven different groups of distance metrics suitable for determining the similarity between density functions according to Che et al. \cite{similaritymetricsstudy}. Thus, the similarity metrics employed for this study are the following: \textit{KL-Divergence (KL)}, \textit{Manhattan Distance (MH)}, \textit{Chebyshev (CBS)}, \textit{Kulsinski (KLS)}, \textit{Cosine (COS)}, \textit{Squared Euclidean (SE)} and \textit{Bhattacharyya (BTC)}. In terms of PCA-kdq, this detector is a modified version of kdqTrees with the purpose of addressing the high dimensionality. The difference between the two is that instead of extracting the data distribution from the original data, we extract it from the projected data, which is computed through PCA. The similarity metrics employed within PCA-kdq are the same as the ones used for kdqTrees. All implementations are publicly available in our replication package\footnote{\url{https://github.com/LorenaPoenaru/concept_drift_detection}}.

\subsection{Evaluation Metrics}
To evaluate the drift detectors, we employ three evaluation metrics, namely the \textit{latency}, the \textit{false positive rate} and the \textit{miss-detection probability}. In our study we use these metrics taking into account our data setup with the purpose of understanding how many testing batches with drift are ignored, how many testing batches without drift are signaled as drift and how many datasets with drift are not reported, respectively.

\textit{\textbf{Latency (L)}}: ranges between 0 and 1 and it shows how late the detector manages to detect the drift. If the detector indicates that there is a drift in the first batch when the drift starts, the latency is 0. Therefore, the latency is 0 if the detector identifies the drift at the batch corresponding to the beginning of concept drift in case of gradual drift and occurrence of concept drift for abrupt drift. 
The formula for the \textit{latency} ($L$) is the following:
\begin{equation}\label{eq1}
L = \frac{k - j}{|B|}; b_j, b_k\in B, 
\end{equation}

\noindent where $b_n$ is the $n$\textsuperscript{th} batch in the list of batches ($B$), $b_j$ is the batch corresponding to the beginning of the concept drift, $b_k$ is the batch detected as drift. This metric takes the value ND (nothing detected) if no drift is detected.

\textit{\textbf{False Positive Rate (FPR)}}: shows the percentage of non-drifted batches detected as drifted batches. If no drift is detected in the data-stream, the metric will output ND (nothing detected). The FPR takes the value 0 if no batch that does not contain drift is signaled as drift and 1 if all batches that do not contain drift are signaled as drift. The formula for the \textit{false positive rate} is the following:\\

\begin{equation}\label{eq2}
FPR = \frac{k^{F}}{|B_{ND}|}; b_k^{F}\in B, 
\end{equation}

\noindent where $b_k^{F}$ is the batch erroneously detected as drift and $B_{ND}$ is the total number of batches without drift out of the total list of batches ($B$).

\textit{\textbf{Miss-Detection Probability (MDP)}}: When evaluating concept drift detectors on synthetic data, it is common to use multiple random seeds of the same dataset to avoid bias. Thus, this metric is only addressed to synthetic data to understand in how many cases the detector managed to identify drift after its occurrence among the 10 random seeds of one dataset, which are referred to as iterations. Since it is a probability, it takes values from 0 to 1, where 0 means that the detector managed to identify drift in all the 10 random seeds of one dataset and 1 means that the detector did not manage to identify any drift in any of the 10 random seeds. The formula for the miss-detection rate is the following: 

\begin{equation}\label{eq4}
MDP = P(L_{(1,...,n)} = ND)
\end{equation}
where n is the number of random seeds, $L_{(1,...,n)}$ is the array corresponding to the latency
\section{Experimental Results}
This section presents the performances achieved by error rate-based (ERB) detectors and data distribution-based (DDB) detectors on both synthetic and real-world data.
\subsection{Synthetic Data}
\subsubsection{Ideal Conditions.} With the scope of addressing the first research question, we conduct the first set of our experiments on synthetically generated data under ideal conditions (no noise or class imbalance added).

We begin our evaluation by assessing the MDP of each detector on each synthetic dataset in case of abrupt drift. Given that all evaluated datasets contain concept drift injected in the process of data generation, we consider that not being able to flag a drift in one iteration of a dataset is an exclusion criteria for the drift detectors in further experiments. Thus, we filter out all detectors with a MDP higher than 0.0 for each dataset. We provide a detailed explanation into which detectors are removed during this step for each dataset together with their corresponding MDP in Table \ref{table:missdetectionratefiltering}. We observe that a high number of DDB detectors achieve an MDP close to 1 in case of AGRAW1 and AGRAW2 datasets, where the categorical data was encoded using one-hot encoding. This shows that the detectors are unable to find differences between the reference data and the upcoming testing data, which could be a consequence of computing the data distribution from a sparse dataset.

\begin{table}[ht]
\centering
\caption{Miss detection probability (MDP) of each excluded detector in case of abrupt drift. In case of the ERB detectors we only show the best MDP of each possible configuration of detector+classifier.}
\begin{tabular}{||c | c | c| c||} 
 \hline
 \textbf{Dataset} & \textbf{Detector Group} & \textbf{Detector} & \textbf{MDP}  \\ [0.5ex] 
 \hline\hline
 SEA & ERB & DDM & 1 \\
 &  & EDDM & 1\\
 & & HDDM\_A  & 0.8\\
  \cline{2-4}
 & DDB & EDE-MW & 1\\
 &  & PCA-kdq & 0.8\\
 \hline
 AGRAW1 & ERB & DDM & 1\\
 &  & EDDM & 1\\
 &  & HDDM\_A & 0.7\\
 \cline{2-4}
 & DDB & EDE-MW & 1\\
 &  & EDE-KS & 0.8\\
 &  & kdqTrees-KL & 1 \\
 &  & kdqTrees-MH &  0.8\\
 &  & kdqTrees-KLS & 1 \\
 &  & kdqTrees-CBS & 1 \\
 &  & kdqTrees-COS & 1 \\
 &  & kdqTrees-SE & 1 \\
 &  & kdqTrees-BTC & 1 \\
 &  & PCA-kdq-MHT & 0.8 \\
 &  & PCA-kdq-CBS & 0.3 \\
 &  & PCA-kdq-COS & 0.7 \\
 &  & PCA-kdq-SE & 0.6 \\
 \hline
 AGRAW2 & ERB & DDM & 1\\
 &  & EDDM & 1\\
 &  & HDDM\_A & 0.9\\
  \cline{2-4}
 & DDB & EDE-MW & 1\\
 &  & EDE-KS & 0.7\\
 &  & PCA-kdq-KL & 0.3\\
 &  & PCA-kdq-MH & 0.3\\
 &  & PCA-kdq-KLS & 0.6\\
 &  & PCA-kdq-CBS & 0.3\\
 &  & PCA-kdq-COS & 0.3\\
 &  & PCA-kdq-SE & 0.3\\
 &  & PCA-kdq-BTC & 0.4\\
 \hline

\end{tabular}
\label{table:missdetectionratefiltering}

{\raggedright Acronyms: KL - KL Divergence Distance, MH - Manhattan Distance, KLS - Kulsinski Distance, COS - Cosine Distance, SE - Squared Euclidean Distance, CBS - Chebyshev Distance, BTC - Bhattacharyya Distance, MW - Mann Whitney statistical test, KS - Kolmogorov Smirnov statistical test \par}
\end{table}

We continue our experiments by assessing the latency and false positive rate of the remaining detectors on each dataset with abrupt drift. In Table \ref{tab:latencyandfprabrupt} we show the results of our findings. The main observation that we can draw from Table \ref{tab:latencyandfprabrupt} is that the error-rate based (ERB) detector ADWIN achieves the lowest latency and false positive rate on all datasets, managing to correctly identify all drifts. Furthermore, its performance is independent of the chosen classifier. When it comes to DDB detectors, we can see that they are in general less precise than the ERB detector ADWIN. Furthermore, there is no general similarity metric or statistical test that achieved the highest performance for all datasets. For both datasets AGRAW1 and AGRAW2, there is no best option in terms of choosing one drift detector, since in all cases there is a compromise between latency and false positive rate. For instance, while KL Divergence minimizes the false positive rate, the Kulsinski and Bhattacharyya distance minimize the latency. Moreover, the DDB detectors can more accurately identify concept drift within the dataset SEA, compared to the datasets AGRAW1 and AGRAW2 datasets.

\begin{table}[ht!]
\caption{Average Latency and FPR of each detector over the 10 iterations for abrupt drift. In \textbf{bold} we show the best performing drift detector for each dataset.}
\begin{tabular}{|r |r |r |r |r  |r |r| r | r | r |r | r | r | r |}

 \cline{4-9}
 \multicolumn{1}{p{0.000000001cm}}{} &
 \multicolumn{1}{c}{} &
  \multicolumn{1}{c}{} &
 \multicolumn{2}{|c}{\textbf{SEA}} &
 \multicolumn{2}{|c|}{\textbf{AGRAW1}} &
 \multicolumn{2}{|c|}{\textbf{AGRAW2}} 
 \\
 
 \cline{2-9}
 \multicolumn{1}{p{0.000000001cm}}{} &
 \multicolumn{2}{|c|}{\textbf{Detector}} &
 \multicolumn{1}{|c|}{\footnotesize{$L$}} &
 \multicolumn{1}{|c|}{\footnotesize{$FPR$}} &
 \multicolumn{1}{|c|}{\footnotesize{$L$}} &
 \multicolumn{1}{|c|}{\footnotesize{$FPR$}} &
 \multicolumn{1}{|c|}{\footnotesize{$L$}} &
 \multicolumn{1}{|c|}{\footnotesize{$FPR$}}  \\
 \cline{2-9}
 \hline
  & \scriptsize{ADWIN}& \scriptsize{*} & \footnotesize{\textbf{0.00}} & \footnotesize{\textbf{0.00}} & \footnotesize{\textbf{0.00}} & \footnotesize{\textbf{0.00}} & \footnotesize{\textbf{0.00}} & \footnotesize{\textbf{0.00}} \\
 
 & \scriptsize{HDDM\_W} & \scriptsize{NB} & \footnotesize{-} & \footnotesize{-} & \footnotesize{\textbf{0.00}} & \footnotesize{\textbf{0.00}} & \footnotesize{0.00} & \footnotesize{1.00}  \\
 &  & \scriptsize{HT} & \footnotesize{-} & \footnotesize{-} & \footnotesize{0.04} & \footnotesize{0.00} & \footnotesize{\textbf{0.00}} & \footnotesize{\textbf{0.00}}  \\
\multirow{1}{0.15em}{\rotatebox[origin=c]{90}{\scriptsize{\textbf{ERB}}}} &  & \scriptsize{ADB} & \footnotesize{-} & \footnotesize{-} & \footnotesize{0.04} & \footnotesize{0.00} & \footnotesize{0.08} & \footnotesize{0.00}  \\
 &  & \scriptsize{XGB} & \footnotesize{-} & \footnotesize{-} & \footnotesize{0.04} & \footnotesize{0.00} & \footnotesize{0.02} & \footnotesize{0.00}  \\
 &  & \scriptsize{LGBM} & \footnotesize{-} & \footnotesize{-} & \footnotesize{0.04} & \footnotesize{0.00} & \footnotesize{0.02} & \footnotesize{0.00}  \\

\hline
\hline

 \multirow{15}{0.15em}{\rotatebox[origin=c]{90}{\scriptsize{\textbf{DDB}}}} & \scriptsize{EDE} & \scriptsize{KS} & \footnotesize{\textbf{0.00}} & \footnotesize{\textbf{0.10}} & \footnotesize{-} & \footnotesize{-} & \footnotesize{-} & \footnotesize{-}\\
  \cline{2-9}
 & \scriptsize{kdqTrees} & \scriptsize{KL} & \footnotesize{0.00} & \footnotesize{0.20} & \footnotesize{-} & \footnotesize{-} & \footnotesize{\textbf{0.16}} & \footnotesize{\textbf{0.15}}   \\
 &  & \scriptsize{MH} & \footnotesize{0.00} & \footnotesize{0.40} & \footnotesize{-} & \footnotesize{-} & \footnotesize{\textbf{0.04}} & \footnotesize{\textbf{0.30}}  \\
 &  & \scriptsize{KLS} & \footnotesize{0.00} & \footnotesize{0.40} & \footnotesize{-} & \footnotesize{-} & \footnotesize{\textbf{0.12}} & \footnotesize{\textbf{0.20}} \\
 &  & \scriptsize{CBS} & \footnotesize{0.00} & \footnotesize{0.20} & \footnotesize{-} & \footnotesize{-} & \footnotesize{\textbf{0.12}} & \footnotesize{\textbf{0.20}} \\
 &  & \scriptsize{COS} & \footnotesize{0.00} & \footnotesize{0.20} & \footnotesize{-} & \footnotesize{-} & \footnotesize{\textbf{0.12}} & \footnotesize{\textbf{0.20}} \\
&  & \scriptsize{SE} & \footnotesize{0.00} & \footnotesize{0.15} & \footnotesize{-} & \footnotesize{-} & \footnotesize{\textbf{0.12}} & \footnotesize{\textbf{0.20}} \\
&  & \scriptsize{BTC} & \footnotesize{\textbf{0.00}} & \footnotesize{\textbf{0.10}} & \footnotesize{-} &\footnotesize{-} & \footnotesize{\textbf{0.12}} & \footnotesize{\textbf{0.20}} \\
  \cline{2-9}

  \cline{2-9}

   \cline{2-9}
   
   & \scriptsize{PCA-kdq} & \scriptsize{KL} & \footnotesize{-} & \footnotesize{-} & \footnotesize{\textbf{0.20}} & \footnotesize{\textbf{0.32}} & \footnotesize{-} & \footnotesize{-}   \\
  &  & \scriptsize{MH} & \footnotesize{0.00} & \footnotesize{0.25} & \footnotesize{-} & \footnotesize{-} & \footnotesize{-} & \footnotesize{-}   \\
  & & \scriptsize{KLS} & \footnotesize{0.00} & \footnotesize{0.25} & \footnotesize{\textbf{0.04}} & \footnotesize{\textbf{0.41}} & \footnotesize{-} & \footnotesize{-}\\
  & & \scriptsize{CBS} & \footnotesize{0.00} & \footnotesize{0.30} & \footnotesize{-} & \footnotesize{-} & \footnotesize{-} & \footnotesize{-} \\
  & & \scriptsize{COS} & \footnotesize{0.00} & \footnotesize{0.25} & \footnotesize{-} & \footnotesize{-} & \footnotesize{-} & \footnotesize{-} \\
  & & \scriptsize{SE} & \footnotesize{0.00} & \footnotesize{0.25} & \footnotesize{-} & \footnotesize{-} & \footnotesize{-} & \footnotesize{-} \\
  & & \scriptsize{BTC} & \footnotesize{0.00} & \footnotesize{0.30} & \footnotesize{\textbf{0.07}} & \footnotesize{\textbf{0.36}} & \footnotesize{-} & \footnotesize{-} \\
  \hline

\end{tabular}

\label{tab:latencyandfprabrupt}

{\raggedright Acronyms: NB- Naive Bayes, HT - Hoeffding Trees, ADB - AdaBoost, XGB - XGBoost, LGBM - LightGBM, KL - KL Divergence Distance, MH - Manhattan Distance, KLS - Kulsinski Distance, COS - Cosine Distance, SE - Squared Euclidean Distance, CBS - Chebyshev Distance, BTC - Bhattacharyya Distance, KS - Kolmogorov Smirnov statistical test \par}
\end{table}

\begin{figure}[t]
\includegraphics[width=0.5\textwidth]{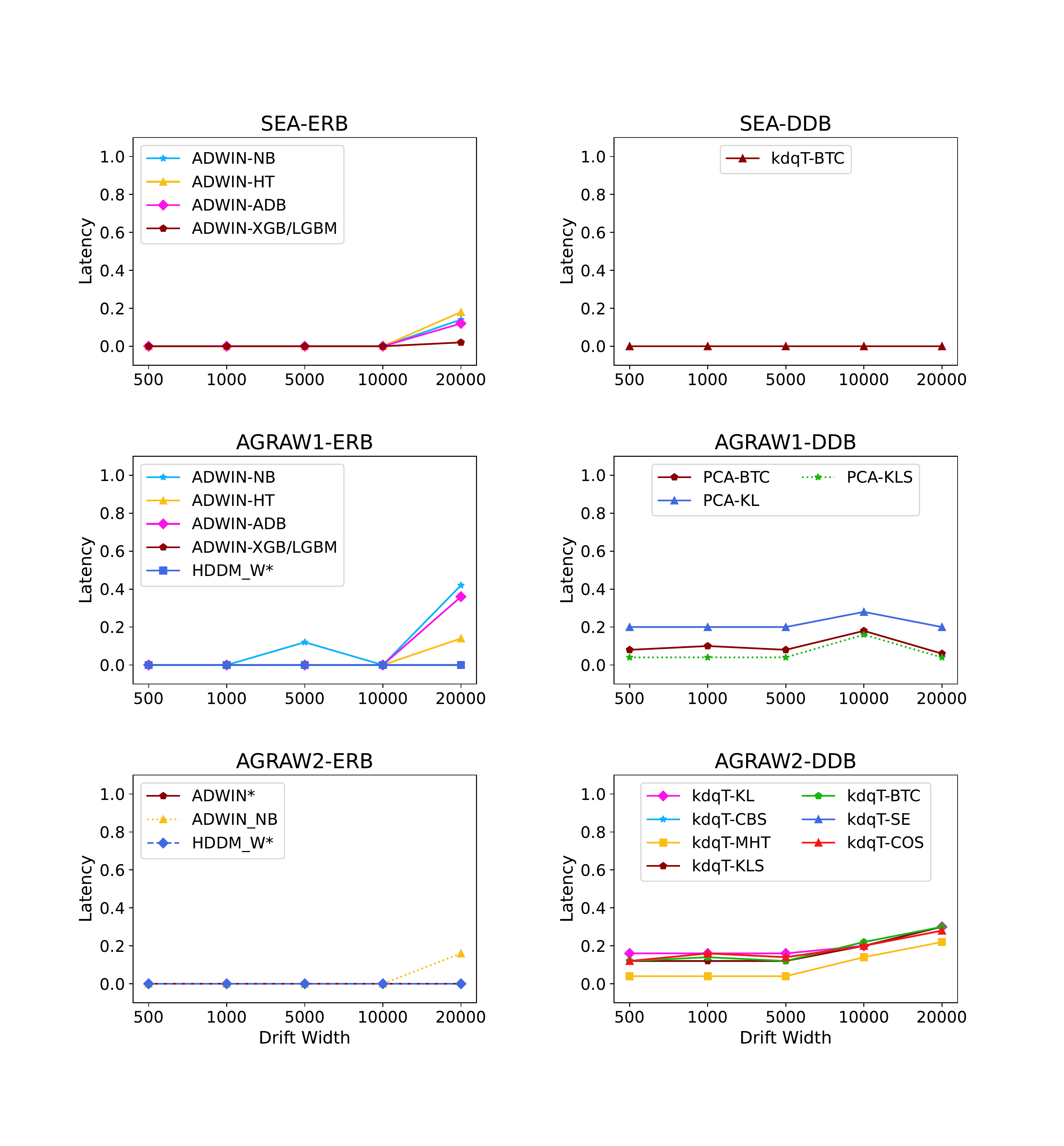}
\caption{Latency of the best performing detectors on different gradual drift width. Each row corresponds to one dataset, SEA, AGRAW1 and AGRAW2. Each column corresponds to the drift detectors type, ERB and DDB.}
\label{figure:drift_width_latency}
\end{figure}

\begin{figure}[t]
\includegraphics[width=0.5\textwidth]{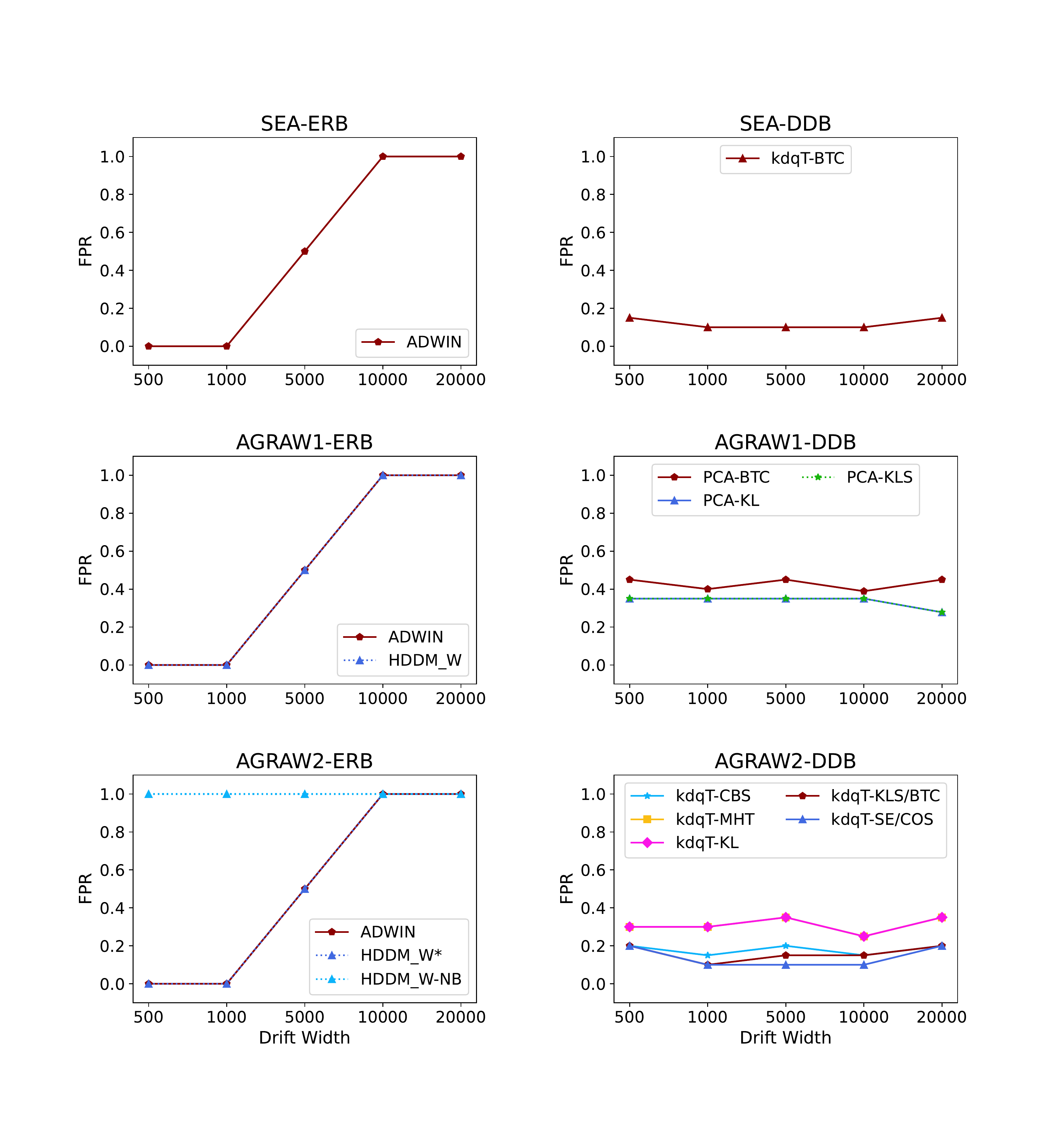}
\caption{False positive rate of the best performing detectors on different gradual drift width. Each row corresponds to one dataset, SEA, AGRAW1 and AGRAW2. Each column corresponds to the drift detectors type, ERB and DDB.}
\label{figure:drift_width_fpr}
\end{figure}

We further assess the precision of both ERB and DDB detectors to identify gradual drift. In this experiment, we solely include the drift best performing drift detectors from the abrupt drift experiment. The reason for this decision is that in real-world settings the type of drift that might occur is unknown and, thereby, we need detectors which work well on both abrupt and gradual drift. We depict the latency and false positive rate of the chosen detectors in Fig.~\ref{figure:drift_width_latency} and Fig.~\ref{figure:drift_width_fpr}, respectively. 

One observation that we can make from Fig.~\ref{figure:drift_width_latency} is that the latency in general not impacted by drift widths lower than 10000 samples. We can notice that the latency of ERB detectors increases slightly for a gradual drift width of 20000 samples. Furthermore, the latency of DDB detectors is overall higher than the latency achieved by the ERB detectors. 

From Fig.~\ref{figure:drift_width_fpr}, we can see that the ERB detectors are severely impacted by higher drift widths, with the false positive rate increasing up to 1.0 for 10000 and 20000 samples. However, the false positive rate of DDB detectors remains relatively stable across the different evaluated drift widths.

\subsubsection{Non-Ideal Conditions} To answer the second research question, we conduct experiments on both abrupt and gradual drift under non-ideal conditions, such as noisy data and class imbalance. In terms of the gradual drift we fix the drift width to 1000 samples, since we noticed from the previous experiment that this is the higher evaluated drift width for which the false positive rate remains 0.0 in case of ERB detectors. However, we observed that the performance of identifying drift in time of ERB and DDB is not affected by the presence of noise. Thus we are not reporting results from this experiment in this section, but we are arguing about the results in the discussions Section. In this experiment we consider the same detectors evaluated on the gradual drift.

\textbf{Class-Imbalance:} One notable outcome of this experiment is the inability of DDB detectors to identify any concept drift when the two classes are imbalanced. This is supported by the increase in miss detection rate, which can be observed in Table~\ref{table:missdetectionrateclassimbalance}. It needs to be mentioned that, similarly to the gradual drift experiment, we solely considered detectors that obtained a miss detection rate equal to 0.0 during the abrupt drift experiment. We can remark from Table~\ref{table:missdetectionrateclassimbalance} that the miss detection rate increases for all detectors, except for the kdqTrees paired with the Manhattan distance when evaluated on the dataset SEA with abrupt drift. However, on the dataset SEA with gradual drift, we can still observe a 0.2 increase of the miss detection rate, showing that this detector was not able to detect any drift in 2 out of 10 random seeds.

\begin{table}[ht]
\centering
\caption{Miss detection probability (MDP) of DDB detector on class imbalance experiment.}
\begin{tabular}{||c | c | c| c||} 
 \hline
 \textbf{Dataset} & \textbf{Detector} & \textbf{MDP Abrupt} & \textbf{MDP Gradual}  \\ [0.5ex] 
 \hline\hline
 SEA & kdqTrees-KL & 0.6 & 0.7 \\
  & kdqTrees-MH & 0.0 & 0.2 \\
  & kdqTrees-KLS & 0.9 & 0.9 \\
  & kdqTrees-CBS & 0.3 & 0.1 \\
  & kdqTrees-COS & 0.3 & 0.3 \\
  & kdqTrees-SE & 0.3 & 0.4 \\
  & kdqTrees-BTC & 0.9 & 0.9 \\
  & PCA-MH & 0.2 & 0.2 \\
  & PCA-KLS & 0.5 & 0.5 \\
  & PCA-CBS & 0.2 & 0.2 \\
  & PCA-COS & 0.2 & 0.2 \\
  & PCA-SE & 0.2 & 0.2 \\
  & PCA-BTC & 0.7 & 0.6 \\
 \hline
 AGRAW1& PCA-KL & 0.5 & 0.5 \\
  & PCA-KLS & 0.5 & 0.6 \\
  & PCA-BTC & 0.6 & 0.6 \\
 \hline
 AGRAW2 & kdqTrees-KL & 0.8 & 0.7\\
  & kdqTrees-MH & 0.6 &  0.7\\
  & kdqTrees-KLS & 0.7 &  0.7\\
  & kdqTrees-CBS & 0.7 &  0.7\\
  & kdqTrees-COS & 0.7 & 0.7 \\
  & kdqTrees-SE & 0.7 & 0.7 \\
  & kdqTrees-BTC & 0.7 & 0.7 \\
 \hline

\end{tabular}
\label{table:missdetectionrateclassimbalance}

{\raggedright Acronyms: KL - KL Divergence Distance, MH - Manhattan Distance, KLS - Kulsinski Distance, COS - Cosine Distance, SE - Squared Euclidean Distance, CBS - Chebyshev Distance, BTC - Bhattacharyya Distance \par}
\end{table}


Since we are dealing with class imbalance, during the experiments with the ERB detectors we initially applied SMOTE~\cite{smote}, which is a commonly used technique that synthetically generates synthetic samples of the minority class. The reason behind this decision is that the classifiers that are paired with the detectors need balanced data to properly learn the behavior of the samples belonging to the two classes. SMOTE was solely applied on the training data in the process of training the classifiers.

\begin{table}[ht!]
\caption{ Latency (L) and False Positive Rate (FPR) for Error Rate-Based detectors on balanced vs imbalanced data. * shows that the results are applicable to all paired classifiers except for the ones presented. - shows that the experiment is not applicable. }
\begin{center}

\hspace*{-0.4cm}
\begin{tabular}{|c |c |c |c | c | c | c | c| c|}

 \cline{2-9}

 \multicolumn{1}{c}{\textbf{}} &
 \multicolumn{1}{|c}{\textbf{Detector}} &
 \multicolumn{1}{|c}{\textbf{Paired}} &
 \multicolumn{2}{|c|}{\textbf{SEA}} &
 \multicolumn{2}{|c|}{\textbf{AGRAW1}} &
 \multicolumn{2}{|c|}{\textbf{AGRAW2}} \\
 \cline{4-9}
 
  \multicolumn{1}{c}{} &
 \multicolumn{1}{|c}{} &
  \multicolumn{1}{|c}{\textbf{with}} &
  \multicolumn{1}{|c|}{\textit{L}} &
 \multicolumn{1}{|c|}{\textit{FPR}} &
  \multicolumn{1}{|c|}{\textit{L}} &
\multicolumn{1}{|c|}{\textit{FPR}} &
  \multicolumn{1}{|c|}{\textit{L}} &
\multicolumn{1}{|c|}{\textit{FPR}} \\
 
 \cline{3-8}
 \hline

 \multirow{4}{0.2em}{\rotatebox[origin=c]{90}{Abrupt}} & \textbf{ADWIN} & * & 0.8 & 0.0 & 0.8 & 0.0 & 0.8 & 0.0 \\
  & & ADB & - & - & - & - & 0.72  & 0.0 \\
  & & HT & - & - & - & - & 0.82 & 0.0 \\
 \cline{2-9}
 & \textbf{HDDM\_W} & * & - & - & 0.0 & 1.0 & 0.0 & 1.0 \\
 \hline
\multirow{3}{0.2em}{\rotatebox[origin=c]{90}{Gradual}} & \textbf{ADWIN} & HT & 0.72 & 0.0 & 0.72 & 0.0 & 0.82 & 0.0\\
& \textbf{ADWIN} & * & 0.8 & 0.0 & 0.8 & 0.0 & 0.8 & 0.0 \\
\cline{2-9}
& \textbf{HDDM\_W} & * & - & - & 0.0 & 1.0 & 0.0 & 1.0 \\
\hline

\end{tabular}

\label{tab:imbalance_abrupt_gradual}
\end{center}

{\raggedright Acronyms: HT - Hoeffding Trees, ADB - AdaBoost \par}
\end{table}

In Table~\ref{tab:imbalance_abrupt_gradual} we show the performances of the two ERB detectors on imbalanced classes. Despite achieving a latency of 0.0, we can notice that the FPR of the HDDM\_W detector significant increased in this setup, signaling every testing batch as a drift batch. Thus, this detector tends to signal a high number of false alarms when used in a real-world setting. When it comes to ADWIN, we can notice that its latency significantly increased compared to the case when the two classes are balanced presented above, but the FPR remains constant at 0.

\subsection{Real-World Data}

This last set of experiments seek to answer RQ3, by understanding how do the analyzed drift detectors perform on real-world data. Here we do not know whether the observed concept drift is abrupt or gradual, but only the position of the drift occurrence. 
\subsubsection{Electricity (ELECT2)} As aforementioned, we assessed the detectors' performances to detect the week in which the 2\textsuperscript{nd} of May 1997 is included and we show the results in Table~\ref{tab:energy}. Here we notice that both ERB and DDB detectors succeed in identifying the exact testing batch which contains the drift. Specifically, the ERB detector called ADWIN managed to exactly identify the drifted batch, independently of the paired classifier. Furthermore, the same results were reported for the DDM classifier paired with Naive Bayes, Hoeffding Trees and AdaBoost. We can further see that using DDM with XGBoost or LightGBM significantly increases its false positive rate from 0.0 to 1.0, enhancing the risk of false alarms. On the other hand, comparable performance was obtained by one DDB detector, namely PCA-kdq using the Kulsinski distance, which also managed to obtain both latency and false positive rate of 0.0. Thereby, for this real-world dataset, DDB detectors managed to achieve comparable good results with ERB detectors.

\begin{table}[ht!]
\caption{Latency (L) and False Positive Rate (FPR) of each detector on ELECT2 dataset. Each detector is paired with either a classifier (for ERB) or a distance/statistical test (for DDB). In \textbf{bold} we show the best performing drift detector(s) from each group. * shows that the results are applicable for any combination and *- shows that results are applicable for any combination except the presented one.}
\begin{center}

\begin{tabular}{|c |c |c |c |c |  }

 \cline{1-5}

 \multicolumn{1}{|c|}{\textbf{Group}} &
 \multicolumn{1}{c}{\textbf{Detector}} &
 \multicolumn{1}{|c|}{\textbf{Paired with}} &
 \multicolumn{1}{|c|}{$L$} &
 \multicolumn{1}{|c|}{$FPR$} \\
 \hline
 \hline
   & \textbf{\small{DDM}}& \textbf{NB, HT, AB} & \textbf{0.0} & \textbf{0.0} \\
   & \small{DDM}& XGB, LGBM & 0.0 & 1.0 \\
 & \small{EDDM}& \small{*} & 0.0 & 1.0 \\
 ERB & \textbf{\small{ADWIN}} & \small{*} & \textbf{0.0 }& \textbf{0.0} \\
 
 & \small{HDDM\_W} & \small{*} & 0.0 & 1.0\\
  & \small{HDDM\_A} & \small{*} & 0.0 & 1.0\\

 \cline{1-5}

 & EDE & * & 0.0 & 1.0\\

  \cline{2-5}
DDB & \small{kdqTrees} & KLS & ND & ND\\
&  & *- & 0.0 & 1.0\\

  \cline{2-5}

  \cline{2-5}

   \cline{2-5}
   
   & \textbf{PCA-kdq} & \textbf{KLS} & \textbf{0.0} &\textbf{ 0.0 } \\
  &  & *- & 0.0 & 1.0 \\

  \hline
  \hline

\end{tabular}

\label{tab:energy}
\end{center}
{\raggedright Acronyms: NB - Naive Bayes,  HT - Hoeffding Trees, 
  AB - AdaBoost, 
  XGB - XGBoost, 
  LGBM - LightGBM, 
  KLS - Kulsinski Distance  \par}
\end{table}

\subsubsection{Airlines} As previously mentioned, in case of this dataset the detectors should detect drift on the evaluation batch corresponding to Friday. In Table~\ref{tab:airlines} we depict the results for both ERB detectors and DDB detectors. Here we can observe that the ERB detectors show a poor performance on the Airlines datasets in terms of latency and false positive rate. Most of the detectors capture the exact moment of drift occurrence, but with the high cost of signaling false alarms. The best false positive rate (0.5) and latency (0.0) was reported by ADWIN paired with the Naive Bayes classifier. The high number of false positives can also be observed in case of most DDB detectors, except for kdqTrees and PCA-kdq paired with the Kulsinski distance, where they did not manage to identify any drift. Thus, both ERB and DDB detectors are affected by false alarms.

\begin{table}[ht!]
\caption{Latency (L) and False Positive Rate (FPR) of each Error Rate-Based (ERB) detector on Airlines dataset. In \textbf{bold} we show the best compromise between the latency and false positive rate. * shows that the results are applicable for any combination and *- shows that results are applicable for any combination except the presented one.}
\begin{center}

\begin{tabular}{|c |c |c |c |c| }

 \cline{1-5}

 \multicolumn{1}{|c}{\textbf{Group}} &
 \multicolumn{1}{|c}{\textbf{Detector}} &
 \multicolumn{1}{|c}{\textbf{Paired with}} &
 \multicolumn{1}{|c|}{$L$} &
 \multicolumn{1}{|c|}{$FPR$} \\
 \hline
 \hline

 ERB & \small{DDM}& NB, HT & ND & 0.5 \\
  && *- & 0.0 & 1.0 \\
   \cline{2-5}
  &\small{EDDM} & NB & ND & 0.5 \\
  & & HT, AB, LGBM & 0.0 & 1.0 \\
  & & XGB & 0.67 & 1.0 \\
 \cline{2-5}
 & \textbf{\small{ADWIN}} & \textbf{NB} & \textbf{0.0} & \textbf{0.5} \\
 & & \small{*-} & 0.0 & 1.0 \\
 
   \cline{2-5}
 & \small{HDDM\_W} & \small{*} & 0.0 & 1.0 \\
  \cline{2-5}
 & \small{HDDM\_A} & \small{*} & 0.0 & 1.0 \\
 \hline
 DDB & EDE & * & 0.0 & 1.0 \\
 \cline{2-5}
  & kdqTree & KLS & ND & ND \\
  & kdqTree & *- & 0.0 & 1.0 \\
  \cline{2-5}
  & PCA-kdq & KLS & ND & ND \\
  & PCA-kdq & *- & 0.0 & 1.0 \\
  
  \hline

\end{tabular}

\label{tab:airlines}
\end{center}
{\raggedright Acronyms: NB - Naive Bayes, HT - Hoeffding Trees, ADB - AdaBoost, XGB - XGBoost, LGBM - LightGBM, KLS - Kulsinski Distance \par}
\end{table}

\section{Discussions}
This section highlights the most important observation that we made during our study regarding the two groups of concept drift detectors, namely the \textit{error rate-based (ERB)} detectors and the \textit{data distribution-based (DDB)} detectors. Therefore, we aim to help practitioners employ the most suitable drift detector according to their data.

\paragraph{ERB detectors proved to be more suitable for datasets including both categorical and numerical features compared to DDB detectors}
One major observation that we can draw from our experiments addressing RQ1 and RQ3 is the fact that DDB detectors achieve higher performance on datasets with solely numerical values, such as SEA and ELECT2, compared to datasets with both numerical and categorical values, such as AGRAW1, AGRAW2 and Airlines. This could be a consequence of the one-hot encoding technique used to transform categorical variables into numerical. This preprocessing technique increases the sparsity of the dataset, since it represents each categorical value as a binary vector. Sparsity usually alters the representation of the data distribution given that the density function is computed using a high number of 0s and 1s~\cite{sparsity}. This impacts the performance of  DDB detectors due to their high dependency on data distributions. However, the ERB detectors do not suffer from this problem, since they rely on the performance of the classifiers, which are robust towards sparse data. 

\paragraph{DDB detectors can achieve high performance solely when the data is scaled} During our experiments we scaled all datasets, such that their values would range in the interval of [0, 1]. Although scaling is a common practice in ML, it is not necessary when using tree-based algorithms, since they are already robust to widely distributed data~\cite{datascaling}. Therefore, we observed that data scaling did not impact the ERB detectors, which rely on ML classifiers' performances. However, we noticed a high impact of unscaled data on the performance of DDB detectors, which were not able to identify any drift. This could be the result of the fact that they solely rely on the data distribution to detect drifts. Having values widely distributed results in a skewed density function, which impacts the ability of similarity metrics to identify significant discrepancies between two data distributions. Furthermore, we experimented with different scaling intervals, but the [0, 1] interval was the most suitable for all the analyzed datasets.

\paragraph{ERB detectors outperform DDB detectors for abrupt and gradual drift with a small drift width, but suffer from a high number of false alarms in case of gradual drift with a large drift width} When conducting experiments for RQ1, we empirically proved that in case of abrupt and small width gradual drift, the ERB drift detectors outperform the DDB drift detectors, achieving a lower latency and a lower false positive rate.  The best performing ERB drift detector overall is ADWIN, which obtained the best latency and false positive rate independently of the chosen dataset or the paired classifier. However, when tested on synthetic data which contains gradual drift with large drift width, the ERB detectors starts signaling multiple false alarms, although the latency is not affected. The same behavior can be noticed when testing the ERB drift detectors on the Airlines dataset, where all detectors suffer from a significantly high false positive rate, which can indicate that this real-world dataset contains a gradual drift. In real-world data the drift type, abrupt or gradual, cannot be controlled. Thus, in a real-world scenarios we should use a detector that is able to identify all types of drifts. Consequently, it is doubtful whether ERB detectors could be employed in practice.

\paragraph{Given the high discrepancy between synthetic and real-world data, there is currently no clear evidence regarding the fact that class imbalance influences the impact of either DDB or ERB detectors} We investigated the effect of class imbalance on concept drift detectors. In case of synthetic data, we noticed that all evaluated detectors suffer from sever performance degradation, even for a small class imbalance ratio of 1:2. However, on the real-world data the detectors behavior was completely different. On the ELECT2 dataset, both ERB and DDB detectors managed to accurately identify concept drift even if the imbalance ratio of the drifted batch was approximately 1:6. However, on the Airlines dataset, both ERB and DDB detectors encountered difficulties when detecting the concept drift in time, although the imbalance ratio of the drifted testing batch was smaller than the one on the synthetic data, namely 1:1.66. This casts doubt on whether the synthetic data manages to mimic the behavior of real-world data when it comes to class imbalanced datasets with concept drift and shows how the performance of detectors on controlled drift position does not generalize to real-world data (RQ3). Therefore, there is no clear evidence of how the class imbalance influences the performance of drift detectors.

\paragraph{When using DDB detectors in practice, multiple similarity distance should be evaluated and in some cases a compromise between latency and false positive rate is required} Another observation that we want to highlight is regarding the DDB detectors. In literature, the most commonly used similarity metric in this detector category is KL Divergence. However, in our experiments for RQ1 and RQ3 we noticed that this similarity metric did not always achieve the lowest latency or false positive rate. When it comes to the synthetic data, the KL Divergence is mostly minimizing the false positive rate, while the Kulsinski Distance or the Bhattacharyya distance minimized the latency. Thereby, when employed in practice, for some datasets a compromise should be made regarding whether the latency should be prioritized over the false positive rate or vice-versa. Furthermore, on the ELECT2 real-world dataset, the PCA-kdq detector paired with the Kulsinski distance achieved the lowest latency and false positive rate. Furthermore, we have not identified any optimal configuration of drift detector + similarity metric that achieved the best performance on all datasets.

\paragraph{The presence of noise does not impact the latency or false positive rate of either ERB or DDB detectors on synthetic data} When answering RQ2, we noticed that the latency and false positive rate of both ERB and DDB detectors are relatively stable against noise. The explanation behind this aspect is that both the reference data and the evaluation data are affected by the same type and percentage of noise. Thus, the differences between the reference data and evaluation data are too small to impact the performance of evaluated drift detectors. Unfortunately, the MOA framework does not have an option to select which parts should be affected by noise or to include different noise percentages in different parts of the data stream. Therefore, we could not investigate the effect of having clean reference data and noisy evaluation data.

\section{Conclusions}
In this paper we have provided an in depth comparison between two categories of drift detectors, the error rate-based drift detectors and the data distribution-based drift detectors under different conditions, synthetic data with ideal conditions, synthetic data with non-ideal conditions and real-world data. For the latter, we have explored multiple similarity metrics and we have observed that some similarity metrics achieved better latency and false positive rate compared to the state-of-the-art KL Divergence on some datasets. Furthermore, we implemented the most popular data distribution based drift detectors and publicly shared them on GitHub and we evaluated the error rate-based drift detectors on three recent and popular classifiers. Additionally, we provided a list of major observations, which aim to serve as guidelines for practitioners that want to include drift detectors to monitor streaming data.

Our observations indicate that the analyzed concept drift detectors are not fully reliable when used as alarming systems. We show empirical evidence of the fact that the error-based drift detectors are signaling false alarms for a high drift width. This questions their ability to detect drifts in environments where the features are slowly changing over time. An example of such situation is the inflation, which does not have immediate impact on the financial features, but it affects them over a longer period of time. When it comes to data distribution-based drift detectors, their performance overall was lower than the error-based drift detectors, although they were not affected by higher drift widths. We observed that in cases of datasets with categorical features, the data distribution-based detectors suffered from high false positive rate. Furthermore, they are also affected by a high miss-detection rate, which can be a consequence of computing the data distribution from a sparse dataset resulted after applying one-hot encoding. This reduces the reliability of drift detectors when used as alarming systems.

In order to advance the field of concept drift detection, we believe that research should focus on developing more data distribution-based detectors. One major limitation of these detectors was the high false positive rate. This might be an indicator that they are too sensitive to small changes in data. However, these small changes might not affect the performance of the ML models. Thus, a promising research direction is exploring which similarity metrics are the least impacted by small changes in data. Moreover, changes in some features might affect the ML models' performances than others. Thereby, another way to improve these drift detectors is to identify the most significant features and monitor drift by computing the data distribution corresponding to only those features. Furthermore, we can explore other ways of encoding categorical data that can reduce the data sparsity, since we noticed that their performance was much lower on datasets where one-hot encoding was employed. When it comes to error rate-based detectors, future research should focus on adapting them to identify gradual drift with a large drift width without signaling false alarms and, thereby, preserving the false positive rate. This study was limited to publicly available implementations of drift detectors. Thereby, we strongly encourage researchers who implement new drift detectors to publicly share their code. Furthermore, in the situation of class imbalance we noticed a strong inconsistency between synthetic and real-world data when it comes to the performance of error rate-based detectors and data distribution-based detectors. Thus, the concept drift research path would benefit from understanding whether the synthetic data is suitable to simulate the behavior of real-world data in case of highly imbalanced classes.

\section*{Acknowledgment}
This work was partially supported by ING through the AI for Fintech Research Lab with TU Delft.

\bibliography{bibliography}
\bibliographystyle{plain}

\end{document}